%% file: aaai25.tex
\documentclass[letterpaper]{article} 
\usepackage{aaai25}  
\usepackage{times}  
\usepackage{helvet}  
\usepackage{courier}  
\usepackage[hyphens]{url}  
\usepackage{graphicx} 
\usepackage{natbib}  
\usepackage{caption} 
\usepackage{algorithm}
\usepackage{newfloat}
\usepackage{listings}
\DeclareCaptionStyle{ruled}{labelfont=normalfont,labelsep=colon,strut=off} 
\floatstyle{ruled}
\newfloat{listing}{tb}{lst}{}
\floatname{listing}{Listing}
\usepackage[utf8]{inputenc} 
\usepackage{booktabs}
\usepackage{amsfonts}
\usepackage{nicefrac}
\usepackage{xcolor}
\usepackage{lipsum}
\usepackage{algpseudocode}
\usepackage{subcaption} 
\usepackage{amssymb}
\usepackage{amsmath}
\usepackage{amsthm}
\usepackage{multirow}
\usepackage{enumitem}
\usepackage{colortbl} 
\usepackage{pifont}
\usepackage{bm}
\usepackage{enumitem}
\usepackage{makecell}

\theoremstyle{plain}
\theoremstyle{definition}

\def\rvx{{\mathbf{x}}}
\def\rvc{{\mathbf{c}}}

\def\ptheta{{\bm{\theta}}}

\newcommand{\coloneqq}{\mathrel{\mathop:}=}
\newcommand{\Input}{\item[\textbf{Input:}]}
\newcommand{\Output}{\item[\textbf{Output:}]}
\algrenewcommand\algorithmiccomment[1]{\hfill \(\triangleright\) #1}

\title{Diffusion-based Semantic Outlier Generation via Nuisance Awareness for Out-of-Distribution Detection}

\author{
    Suhee Yoon\textsuperscript{\rm 1}\equalcontrib,
    Sanghyu Yoon\textsuperscript{\rm 1}\equalcontrib, 
    Ye Seul Sim\textsuperscript{\rm 1},\\
    Sungik Choi\textsuperscript{\rm 1},
    Kyungeun Lee\textsuperscript{\rm 1},
    Hye-Seung Cho\textsuperscript{\rm 1},
    Hankook Lee\textsuperscript{\rm 2}\thanks{Corresponding Authors.},
    Woohyung Lim\textsuperscript{\rm 1}\textsuperscript{\textdagger}
}
\affiliations{
    \textsuperscript{\rm 1}LG AI Research \\
    \textsuperscript{\rm 2}Sungkyunkwan University \\
    \{suhee.yoon, sanghyu.yoon, ysl.sim, sungik.choi, kyungeun.lee, hs.cho, w.lim\}@lgresearch.ai \\ hankook.lee@skku.edu
}

\begin{document}

\maketitle

\begin{abstract}
Out-of-distribution (OOD) detection, which determines whether a given sample is part of the in-distribution (ID), has recently shown promising results through training with synthetic OOD datasets. Nonetheless, existing methods often produce outliers that are considerably distant from the ID, showing limited efficacy for capturing subtle distinctions between ID and OOD. To address these issues, we propose a novel framework, \textbf{Semantic Outlier generation via Nuisance Awareness (SONA)}, which notably produces challenging outliers by directly leveraging pixel-space ID samples through diffusion models. Our approach incorporates \textit{SONA guidance}, providing separate control over semantic and nuisance regions of ID samples. Thereby, the generated outliers achieve two crucial properties: (i) they present explicit semantic-discrepant information, while (ii) maintaining various levels of nuisance resemblance with ID. Furthermore, the improved OOD detector training with SONA outliers facilitates learning with a focus on semantic distinctions. Extensive experiments demonstrate the effectiveness of our framework, achieving an impressive AUROC of 88\% on near-OOD datasets, which surpasses the performance of baseline methods by a significant margin of approximately 6\%.
\end{abstract}

\input{1_introduction}
\input{2_preliminaries}

\input{3_methodology}

\input{4_experiments}
\input{5_relatedwork}
\input{6_conclusion}

\bibliography{aaai25}

\appendix
\input{7_supplement} 

\end{document}

%% file: 1_introduction.tex
\section{Introduction} \label{sec:1}

\emph{Out-of-distribution (OOD) detection} is a fundamental machine learning task which aims to detect whether a given sample is drawn from the in-distribution (ID) or not. Among a number of OOD detection methods \cite{Kimin2018Training,Si2018Open,Pdtra2018Discri,Andrey2018Pred,Dan2017A}, one promising approach is to learn a detector using auxiliary OOD samples, as pioneered by Outlier Exposure \cite[OE;][]{Hendrycks2019Deep}. This makes learning relatively easier since such outliers can provide practical heuristics for OOD detection. Although this approach has achieved outstanding performance in the recent literature \cite{Deepak2020Explore, goyal2020DROCC, Jin2021Seman, Cai2022Pertu, Jin2023Mixture}, they suffers from the outlier acquisition problem since determining whether a sample qualifies as an outlier is difficult.

\begin{figure}[t]
    \centering
	\includegraphics[width=0.8\linewidth]{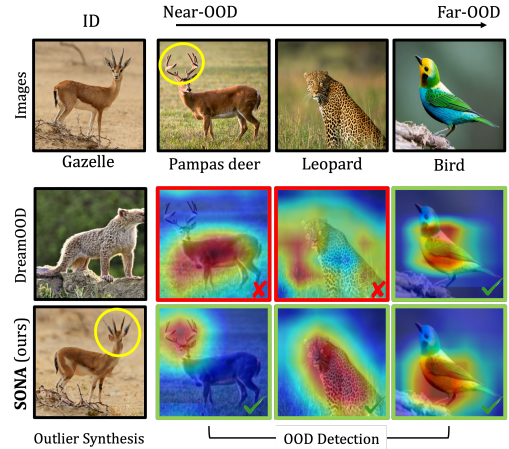}
	\caption{\textbf{OOD examples with Grad-CAM highlighting crucial regions for OOD detection.} DreamOOD, a recent baseline, succeed in far-OOD detection, yet near-OOD cases pose challenge as crucial semantic region become more focused. SONA, however, allows the detector to capture these subtle semantic distinctions.}

 \label{int:grad_cam}
 \vspace{-0.2in}
\end{figure}

\begin{figure}[t]
    \centering
	\includegraphics[width=1\linewidth]{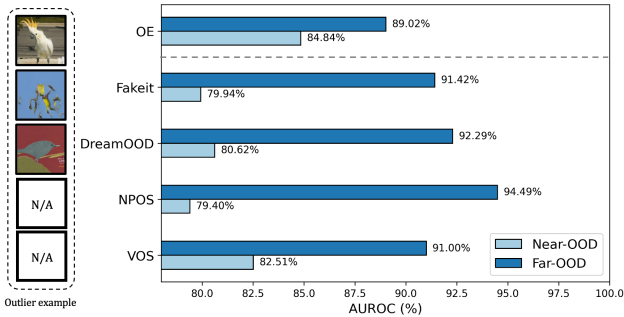}
     \vspace{-0.25in}
	\caption{\textbf{Performance comparison of Near- vs Far-OOD detection with auxiliary outliers.} Only OE utilizes real outliers; the rest methods use synthetic. Outliers for the \emph{junco} bird species (ID) are shown on the left.}
    \vspace{-0.2in}
 \label{motive}
\end{figure}

To mitigate the issue, recent research has explored synthesizing outliers to help the model to learn a precise decision boundary between ID and OOD. For example, VOS \cite{Du2022VOS} synthesizes outliers from the low-likelihood region within a latent space under a distributional assumption (\emph{e.g.}, Gaussian) regarding the ID latent variables. However, this assumption is often inadequate, leading to a failure in capturing the informative features for OOD detection, such as class-specific object details \cite{bai2024id}. Another recently emerging direction is to synthesize outliers within the pixel-space via diffusion models \cite{mirzaei2022fake, du2024dream}, which provides not only high-resolution samples but also visual interpretability. However, the existing works often suffers from volatility in the OOD detection performance as the quality of outliers heavily relies on generation targets (\emph{e.g.,} OOD prompts in DreamOOD \cite{du2024dream} or blurry images in Fakeit \cite{mirzaei2022fake}). In particular, DreamOOD intensifies this volatility by entirely depending on OOD (label/text) prompts since it starts from random noises without pixel-space information in ID images. Consequently, these approaches generate less challenging outliers and exhibit limited efficacy in capturing subtle semantic distinctions (see Figure~\ref{int:grad_cam}). This inadequacy is evident in the significant performance drop on Near-OOD detection, as illustrated in Figure~\ref{motive}.

The key observation underlying this performance drop is that the detectors tend to overlook semantic distinctions required for a wide range of OOD detection scenarios, near-to-far, as shown in Figure~\ref{int:grad_cam}. For instance, while a gazelle image is easily distinguishable from a bird, existing models are often confused with a leopard image due to their similar background, grassland. This challenge intensifies when considering a deer, which shares a body structure with a gazelle, making the delicate feature detection like the antlers even more difficult. As highlighted by \citet{wiles2022fine}, such semantics variations are crucial and commonly exist in datasets inspired by real-world scenarios. Hence, it is necessary to account for the various semantic and nuisance levels in pixel-space ID samples, which retain the most comprehensive information.

\textbf{Contribution.} In this paper, we introduce a novel framework, \textbf{Semantic Outlier Generation via Nuisance Awareness (SONA)}, that notably produces challenging outliers by directly leveraging pixel-space ID samples through diffusion models. In particular, we propose \textit{SONA guidance}, accomplishing two crucial properties for effective outliers: (i) presenting explicit semantic-discrepant information, while (ii) maintaining nuisance resemblance with ID.

Concretely, we first construct the underlying framework of input deformation that decides the extent of original input manipulation and drives it towards the desired direction (Section~\ref{sec:3.1}). To enhance the differentiated impact on semantic and nuisance region, we present a novel approach for identifying non-overlapping regions for each sample (Section~\ref{sec:3.2}). Following this, our simple yet effective method, \textit{SONA guidance}, induces discrete directions and degrees of transformation in each region (Section~\ref{sec:3.3}). Lastly, we introduce an OOD detector training loss designed to enhance the SONA outliers utilization (Section~\ref{sec:3.4})

Through extensive experiments, we demonstrate the effectiveness of the SONA framework not only on widely used far-OOD, but also on the more challenging near-OOD scenarios. Given \textit{ImageNet} as the ID dataset, the SONA framework outperforms the prior outlier synthesis baselines. Moreover, our method remains competitive, avoiding the severe variations seen in existing works that heavily depend on specific generation targets.

Our contributions are summarized as follows:
\begin{itemize}
    \item We propose \textbf{SONA}, an novel outlier generation framework that enhances OOD detectors to capture a wide range of OOD samples, with a particular focus on Near-OOD settings.
    \item Our \emph{SONA guidance} induces fine-grained control over semantic and nuisance regions on pixel-space ID samples using diffusion models. This approach enables to produce effective outliers that closely resemble the ID in terms of nuisance while incorporating semantic-discrepant information.
    \item SONA successfully empowers the detector to capture the subtle semantic discrepancies by showing impressive AUROC of 88.5\% on Near-OOD tasks (Table~\ref{tab:main1}).
\end{itemize}

\begin{figure*}[!t]
\label{fig:overview}
    \centering
	\includegraphics[width=0.8\linewidth]{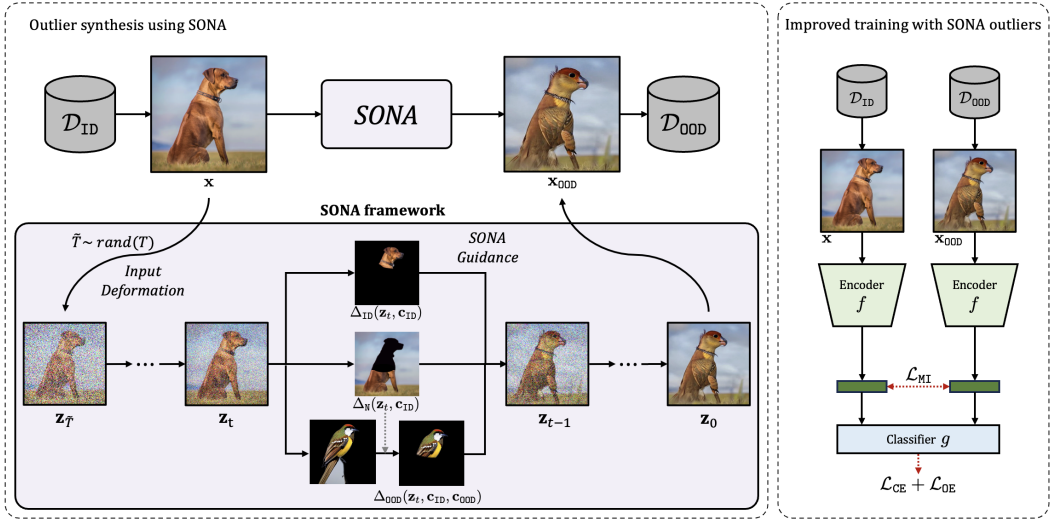}
	\caption{\textbf{Overview of SONA framework.} The process begins with $\mathbf{z}_{\tilde{T}}$, a noisy latent variable with a randomly chosen $\tilde{T}$, undergoes denoising by \emph{SONA guidance}. This guidance strategically introduces semantic discrepancies while maintaining varying degrees of nuisance resemblance across all $\tilde{T}$. The resulting $\mathbf{{x}}_{\texttt{OOD}}$ are used to train the classifier with their source $\mathbf{x}$, focusing on discerning semantic differences.}
 \label{3_overview}
\end{figure*}

%% file: 2_preliminaries.tex
\section{Preliminaries} 
\label{sec:2}

\subsubsection{Out-of-Distribution Detection.} The task of OOD detection aims to identify whether a given input $\mathbf{x}$ is drawn from the training distribution or not. Let $\mathcal{D}_\texttt{ID}=\{\mathbf{x}^{(i)}, {y}^{(i)}\}$ be a training distribution over $\mathcal{X}\times\mathcal{Y}$, where $\mathcal{X}$ denotes the input space and $\mathcal{Y} = \{1, \ldots, C\}$ denotes the label space with $C$ classes.

A feature encoder $f: \mathcal{X} \rightarrow \mathcal{Z}$ is trained on this distribution to map inputs to a feature space $\mathcal{Z}$. Subsequently, a classifier $g: \mathcal{Z} \rightarrow \mathcal{Y}$ is trained to predict the class labels. The trained feature encoder and classifier are then used to develop a scalar function $S_{f,g}: \mathcal{X} \rightarrow \mathbb{R}$, which provides a confidence score to determine if an input $\mathbf{x}$ is within the training distribution (\emph{i.e.}, $S_{f,g}(\mathbf{x}) \leq \kappa$) or not (\emph{i.e.}, $S_{f,g}(\mathbf{x}) > \kappa$), where $\kappa$ is a predefined hyperparameter.

\subsubsection{Conditional Diffusion Models (CDMs).} CDMs have recently shown promising advances in image generation by incorporating specific conditions $\mathbf{c}$ (\emph{e.g.}, class labels or text prompts) into diffusion process. In particular, Classifier-free Diffusion Guidance \cite[CFG;][]{ho2022classifier} represents a simple yet effective approach of CDM, eliminating the need for a separate classifier. During CFG training, they randomly drop the condition with an unconditional probability, optimizing the reverse process parameter $\ptheta$ with the following objective:
\begin{equation*}
    \mathcal{L}(\ptheta)=\mathbb{E}_{(\rvx_{0},\rvc) \sim \mathcal{D}, t\sim \mathcal{U} \left \{1,\ldots,T\right\}, \epsilon \sim \mathcal{N}(\bm{0},\mathbf{I})} \left[\left\| \epsilon_{\ptheta}(\rvx_{t}, \rvc)-\epsilon \right\|^{2}_{2} \right].
\end{equation*}
In the sampling phase, the noise prediction $\tilde{\epsilon}_{\ptheta}(\rvx_{t},\rvc)$ can be expressed as:
\begin{align}
    \tilde{\epsilon}_{\ptheta}(\rvx_{t}, \rvc) 
    &= \epsilon_{\ptheta}(\rvx_{t}) + s\cdot(\epsilon_{\ptheta}(\rvx_{t},\rvc) - \epsilon_{\ptheta}(\rvx_{t})) \nonumber \\
    &= \epsilon_{\ptheta}(\rvx_{t}) + s\cdot\psi(\rvx_{t}, \mathbf{c}),
\label{cfg_sampling}
\end{align}
where $s$ is the guidance scale and $\psi(\mathbf{x}_t, \mathbf{c})$ denotes the difference between conditional and unconditional predictions. While previous outlier generation methods utilized Stable Diffusion \cite{rombach2022high} for its strong performance in generating high-quality images, these methods face challenges due to performance variations caused by the high sensitivity to the conditions chosen from the CLIP text encoder \cite{radford2021learning}, highlighting the need to address this issue.

%% file: 3_methodology.tex
\vspace{-6pt}
\section{Method} \label{sec:3} 
We introduce \emph{Semantic Outlier generation via Nuisance Awareness (SONA)}, a novel and effective outlier synthesizing framework covering Near-to-Far OOD detection scenarios. Our key idea is to directly incorporate full pixel-space ID images into CDM by specifying semantic and nuisance regions. This enables our framework to generate semantic-discrepant outliers with resembling the nuisance of ID samples, which has superiority over prior synthetic outlier-based training methods, especially in Near-OOD detection tasks.

\subsection{Overview}
Our framework begins by introducing the underlying structure of input deformation and guides it to a new desired direction (Section~\ref{sec:3.1}). Following this, we specify semantic and nuisance non-overlapping regions for each sample (Section~\ref{sec:3.2}). Based on these regions, we propose our new region-specific guidance, \textit{SONA Guidance}, denoted by $\Delta_\texttt{SONA}$ (Section~\ref{sec:3.3}). $\Delta_\texttt{SONA}$ allows a diffusion model to deform the original semantic more intensively while remaining the nuisances. The modified noise prediction with our SONA guidance can be expressed as follows:
\begin{equation}
\label{eq:modi_noise}
\tilde{\epsilon}_{\ptheta}(\mathbf{z}_t,  \mathbf{c}_\texttt{ID}, \mathbf{c}_\texttt{OOD}) := \epsilon_{\ptheta}(\mathbf{z}_t) + s\cdot\Delta_\texttt{SONA}(\mathbf{z}_t, \mathbf{c}_\texttt{ID}, \mathbf{c}_\texttt{OOD}),
\end{equation}
where $\mathbf{c}_\texttt{ID}$ and $\mathbf{c}_\texttt{OOD}$ are conditions obtained by ID and OOD labels, respectively, and the latter can be sampled from a broad range of text conditions that does not include in ID. Lastly, the advanced OOD detector training method with SONA outliers is explained on (Section~\ref{sec:3.4}). The comprehensive overview of our framework is illustrated in Figure~\ref{3_overview}.

\vspace{-2pt}
\subsection{Input Deformation for Outlier Synthesis} \label{sec:3.1}

We here describe our underlying framework for input deformation that transforms the original ID sample $\mathbf{x} \in \mathcal{D}_{\texttt{ID}}$ into an outlier. After encoding $\mathbf{x}$ into its latent representation $\mathbf{z}_0$, the deformation process starts by performing an incomplete diffusion process from the clean latent $\mathbf{z}_0$ to a noisy latent $\mathbf{z}_{\tilde{T}}$ where $\tilde{T}\sim\mathcal{U}(1, T)$ is an early stop timestep. By stopping the process earlier, we obtain the noisy latent $\mathbf{z}_{\tilde{T}}$ with more corruption effects in the semantically important areas. This is because Gaussian noise exhibits a uniform spectral density, which makes semantic components more susceptible to perturbations than nuisance~\cite{Wyatt2022Anoddpm, lee2023multi, Wang2020High}. Hereby, denoising from the obtained $\mathbf{z}_{\tilde{T}}$ with the new conditional guidance is supposed to lead to more deformation effect on the semantic rather than nuisance. 

Nevertheless, these denoised samples are not sufficiently qualified as outliers, due to their limited property of varying $\tilde{T}$. Stopping too early $\tilde{T}$ fails to initiate semantic changes, leaving the sample almost identical to the ID and potentially confusing the OOD detector. Conversely, a large $\tilde{T}$ leads to abrupt changes in both semantic and nuisance, resulting in overly distant outliers. We solve this sensitivity issue and improve the quality of the transformation by specifying semantic and nuisance region  (Section ~\ref{sec:3.2}) and proposing an effective guidance term $\Delta_{\texttt{SONA}}$ (Section ~\ref{sec:3.3}).

\vspace{-2pt}
\subsection{Semantic and Nuisance Region Masking} \label{sec:3.2}
Our key strategy is to robustly control the changes in both semantic and nuisance information for any $\tilde{T}$. For this, we identify two non-overlapping regions, the semantic region $M_\texttt{S}$ and the nuisance region $M_\texttt{N}$. We accomplish this by utilizing $\psi$, defined as the difference between the conditional and unconditional noise estimates, \emph{i.e.}, $\psi(\mathbf{z}_{t}, \mathbf{c}) = \epsilon_{\ptheta}(\mathbf{z}_t,\rvc)-\epsilon_\ptheta(\mathbf{z}_t)$.

The \emph{semantic region} $M_\mathtt{S}$ represents the area that captures the unique semantic information of the given context $\mathbf{c}$, which is not shared by other contexts. It corresponds to the portion where the absolute difference between the conditional and unconditional noise estimates is large. As proven in prior work~\cite{Brack2024SEGA}, the top 1-5\% of $|\psi|$ values effectively capture semantic information. Therefore, $M_\mathtt{S}$ can be defined by the percentile threshold $\eta_\lambda$, representing the $\lambda$-th percentile of $|\psi|$.

The \emph{nuisance region} $M_\mathtt{N}$ is a less relevant portion to the context $\mathbf{c}$, as indicated by the lowest absolute differences. Since $M_\mathtt{N}$ contains more generic and redundant information commonly shared with other random contexts, it should be given less importance during the OOD detection process. We determine $M_\mathtt{N}$ using the lowest 1-5\% as well, without employing an additional nuisance threshold hyperparameter. Formally, the semantic and nuisance region masks can be written as follows:
\begin{equation}
\begin{aligned}
M_\texttt{S}(\mathbf{z}_t,\mathbf{c}) &= \begin{cases}
1 & \text{if } |\psi(\mathbf{z}_t, \mathbf{c})| \geq \eta_\lambda, \\
0 & \text{otherwise,}
\end{cases} \\
M_\texttt{N}(\mathbf{z}_t,\mathbf{c}) &= \begin{cases}
1 & \text{if } |\psi(\mathbf{z}_t, \mathbf{c})| < \eta_{1-\lambda}, \\
0 & \text{otherwise.}
\end{cases}
\end{aligned}
\end{equation}
\begin{figure}[!t]
  \centering
  \includegraphics[width=\linewidth]{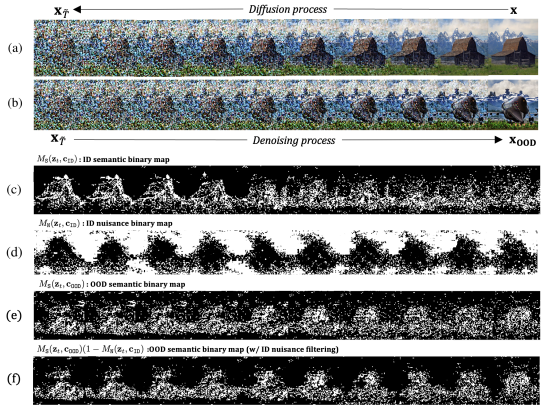} 
  \caption{\textbf{Illustration of the denoising process with SONA guidance.} (a) The diffusion process of an ID image with the original label \textbf{barn} up to $\tilde{T}=35$. (b) The denoising process from timestep $\tilde{T}=35$ to 0 with SONA guidance using the OOD label \textbf{airliner}. (c), (d) and (e) show the ID semantic, ID nuisance, and OOD semantic region mask, respectively, at $\lambda=0.2$. (f) The final OOD semantic region mask obtained by filtering out the intersecting areas between $M_\texttt{S}(\mathbf{z}_t,\mathbf{c}_\texttt{OOD})$ and $M_\texttt{N}(\mathbf{z}_t,\mathbf{c}_\texttt{ID})$.}
  \vspace{-0.2in}
  \label{3.3_denoising}
\end{figure}

\subsection{SONA Guidance} \label{sec:3.3}
In this section, we introduce \textit{SONA guidance}, a novel approach that enables fine-grained control on each semantic and nuisance region for outlier generation. Our method draws inspiration from recent advances in image editing, which aims to completely replace target semantics. However, as they are not inherently designed for OOD detection, their application has shown limited impact in this context (see Appendix). We propose a developed approach adequate for outlier generation, allowing precise control over both $M_\mathtt{S}$ and $M_\mathtt{N}$ to prioritize semantic differences. The SONA guidance term, $\Delta_\texttt{SONA}$, is composed of three components as follows, each of which is described in the following paragraphs:
\begin{equation}
\Delta_\texttt{SONA}:=\Delta_\texttt{ID}+\Delta_\texttt{N}+\Delta_\texttt{OOD}.
\end{equation}

\subsubsection{Removal of ID semantic information.}
During the denoising process, $\Delta_\texttt{ID}$ removes the ID semantic, $\mathbf{c}_\texttt{ID}$, targeting on $M_\texttt{S}$. For this, we change the direction of semantic enhancement in the opposite way as written below. Figure~\ref{3.3_denoising} (c) shows that $M_\texttt{S}$ retains a substantial amount of $\mathbf{c}_\texttt{ID}$ semantics during the initial stages of the denoising process, but by the end of the process, most of the semantic information gradually disappears. This approach effectively mitigates the issue of the original image being restored when the stop timestep $\tilde{T}$ is chosen too early.
\begin{equation}
\Delta_\texttt{ID}(\mathbf{z}_t,\mathbf{c}_\texttt{ID}):=-M_\texttt{S}(\mathbf{z}_t,\mathbf{c}_\texttt{ID})\odot\psi(\mathbf{z}_t,\mathbf{c}_\texttt{ID}).
\end{equation}

\subsubsection{Preservation of ID nuisance information.}
To encourage the detector to focus on changes within the semantic regions, $\Delta_\texttt{N}$ is designed to retain a relative amount of nuisance information (Figure~\ref{3.3_denoising} (d)). By guiding $M_\texttt{N}$ in the direction of $\psi(\mathbf{z}_t,\mathbf{c}_\texttt{ID})$ as written below, we can maintain varying degrees of nuisance information, depending on $\tilde{T}$. Therefore, our method gains more advantage over semantic editing methodologies that aim for explicit nuisance preservation, as our method generates a diverse set of outliers with different levels of nuisance retention.
\begin{equation}
\Delta_\texttt{N}(\mathbf{z}_t,\mathbf{c}_\texttt{ID}):=M_\texttt{N}(\mathbf{z}_t,\mathbf{c}_\texttt{ID})\odot\psi(\mathbf{z}_t,\mathbf{c}_\texttt{ID}).
\end{equation}

\subsubsection{Addition of OOD semantic information.} 
$\Delta_\texttt{OOD}$ serves as another component that induces semantic-discrepant outlier generation by corrupting with new semantic of $\mathbf{c}_\texttt{OOD}$. However, as shown in Figure~\ref{3.3_denoising} (e), $M_\texttt{S}(\mathbf{z}_t,\mathbf{c}_\texttt{OOD})$ slightly extends beyond the original semantic region. To improve the preservation of nuisance and induce the corruption on the semantic region, we further filter out the intersecting parts between ood semantic region $M_\texttt{S}(\mathbf{z}_t,\mathbf{c}_\texttt{OOD})$ and the nuisance region $M_\mathtt{N}(\mathbf{z}_t,\mathbf{c}_\texttt{ID})$ (Figure~\ref{3.3_denoising} (f)). By intentionally rectifying the areas where $\psi(\mathbf{z}_{t},\mathbf{c}_\texttt{OOD})$ has an effect, we ensure that the influence of nuisance attributes remains significant even when $\tilde{T}$ chosen as later timestep.
\begin{equation}
\begin{aligned}
& \Delta_\texttt{OOD}(\mathbf{z}_t,\mathbf{c}_\texttt{ID},\mathbf{c}_\texttt{OOD})  \\
& := M_\texttt{S}(\mathbf{z}_t,\mathbf{c}_\texttt{OOD})\odot(1-M_\mathtt{N}(\mathbf{z}_t,\mathbf{c}_\texttt{ID}))
 \odot\psi(\mathbf{z}_t,\mathbf{c}_\texttt{OOD}).
\end{aligned}
\end{equation}
Finally, we obtain the outliers by denoising $\mathbf{z}_t$ for every timestep $t = \tilde{T}, \dots, 1$ with our SONA guidance. To generate the pixel-space outlier image $\mathbf{x}_\texttt{OOD}$, we pass the denoised latent representation $\mathbf{z}_0$ through the decoder of the diffusion model. A detailed pseudo-code implementation of our framework can be found in the Appendix.

\subsection{OOD Detection with SONA Outliers} \label{sec:3.4}
The generated SONA outliers are used to preciously regularize the classifier, with a focus on semantic aspects. Given $(\mathbf{x},y)\in\mathcal{D}_{\texttt{ID}}$ and $\mathbf{x}_\texttt{OOD}\in\mathcal{D}_{\texttt{OOD}} $, our training objective with SONA is formulated as:
\begin{align}
\mathcal{L} = \: & \mathbb{E}_{(\mathbf{x}, y) \sim D_{\texttt{ID}}} \left[ \mathcal{L}_{\texttt{CE}} (g(f(\mathbf{x})), y) \right] \nonumber \\
& + \beta \mathbb{E}_{\mathbf{x}_{\texttt{OOD}} \sim D_{\texttt{OOD}}} \left[ \mathcal{L}_{\texttt{OE}} (g(f(\mathbf{x}_{\texttt{OOD}}))) \right] \nonumber \\
& +  \mathcal{L}_\texttt{MI}(f(\mathbf{x}), f(\mathbf{x}_\texttt{OOD}))
\end{align}
In the above objective, $\mathcal{L}_\texttt{CE}$ is a cross-entropy loss that compels ID samples to discriminate classes using labels $y$. $\mathcal{L}_\texttt{OE}$ encourages the separation of SONA outliers from ID by inducing their predictions to a uniform distribution, which can be expressed as $-\frac{1}{C} \sum_{c=1}^{C} \text{softmax}_{c}(f(\mathbf{x}_\texttt{OOD}))$. Moreover, we introduce an additional loss term, $\mathcal{L}_{\texttt{MI}}$, which minimizes the mutual information between SONA outliers and their source ID samples. This is achieved through the Kullback-Leibler divergence between the joint distribution and the product of marginal distributions:
\begin{align}
\mathcal{L}_{\texttt{MI}} &= \text{MI}\left( f(\mathbf{x}); f(\mathbf{x}_\texttt{ood}) \right) \nonumber\\
&= \text{KL}\left(P(f(\mathbf{x}), f(\mathbf{x}_\texttt{ood})) \middle| P(f(\mathbf{x}))P(f(\mathbf{x}_\texttt{ood})) \right)
\end{align}
Specifically, we implement this by minimizing the Contrastive Log-ratio Upper Bound \cite[CLUB;][]{Cheng2020Club} at the feature extraction layer, where nuisances are significantly reduced. This enables the classifier to directly compare and learn the semantic differences between ID samples and their corresponding SONA outliers.

During the test phase, we utilize the energy score \cite{Liu2020Energy}, which effectively addresses overconfidence issues in OOD detection by assigning lower energy values to ID samples and higher energy values to OOD samples.

\begin{table*}[htbp]
\centering
\setlength{\tabcolsep}{3pt}
\renewcommand{\arraystretch}{0.85}
\small
\begin{tabular}{@{}lcccccccc@{}}
\toprule
\multicolumn{1}{c}{\multirow{3}{*}{Method}} & \multicolumn{7}{c}{AUROC} & \multirow{3}{*}{ID ACC} \\ \cmidrule(lr){2-8}
\multicolumn{1}{c}{} & \multicolumn{3}{c}{Near-OOD} & \multicolumn{4}{c}{Far-OOD} &  \\ \cmidrule(lr){2-8}
\multicolumn{1}{c}{} & SSB-hard & NINCO & Avg & iNaturalist & Texture & OpenImage-O & Avg &  \\ \midrule
\multicolumn{9}{l}{\textit{\textbf{Post-hoc}}} \\ \midrule
\multicolumn{1}{l|}{OpenMax} & 77.53 & 83.01 & \multicolumn{1}{c|}{80.27} & 92.32 & 90.21 & 88.07 & \multicolumn{1}{c|}{90.20} & 86.37 \\
\multicolumn{1}{l|}{MSP} & 80.38 & 86.29 & \multicolumn{1}{c|}{83.34} & 92.80 & 88.36 & 89.24 & \multicolumn{1}{c|}{90.13} & 86.37 \\
\multicolumn{1}{l|}{ODIN} & 77.19 & 83.34 & \multicolumn{1}{c|}{80.27} & 94.37 & 90.65 & 90.11 & \multicolumn{1}{c|}{91.71} & 86.37 \\
\multicolumn{1}{l|}{EBO} & 79.83 & 86.17 & \multicolumn{1}{c|}{82.50} & 92.55 & 90.79 & 89.23 & \multicolumn{1}{c|}{90.86} & 86.37 \\
\multicolumn{1}{l|}{OpenGAN} & 55.08 & 69.49 & \multicolumn{1}{c|}{59.79} & 75.32 & 70.58 & 73.54 & \multicolumn{1}{c|}{73.15} & 86.37 \\
\multicolumn{1}{l|}{ReAct} & 78.97 & 84.76 & \multicolumn{1}{c|}{81.87} & 93.65 & 92.86 & 90.40 & \multicolumn{1}{c|}{92.31} & 86.37 \\
\multicolumn{1}{l|}{KNN} & 77.03 & 86.10 & \multicolumn{1}{c|}{81.57} & 93.99 & 95.29 & 90.19 & \multicolumn{1}{c|}{93.16} & 86.37 \\
\multicolumn{1}{l|}{DICE} & 79.06 & 84.49 & \multicolumn{1}{c|}{81.78} & 91.81 & 91.53 & 89.06 & \multicolumn{1}{c|}{90.80} & 86.37 \\ 
\midrule
\multicolumn{9}{l}{\textit{\textbf{Training w/ Real-World Outlier}}} \\ \midrule
\multicolumn{1}{l|}{OE} & 82.34 & 87.35 & \multicolumn{1}{c|}{84.84} & 90.30 & 87.76 & 89.01 & \multicolumn{1}{c|}{89.02} & 85.82 \\
\multicolumn{1}{l|}{MCD} & 81.51 & 85.74 & \multicolumn{1}{c|}{83.62} & 90.83 & 86.87 & 89.12 & \multicolumn{1}{c|}{88.94} & 86.12 \\
\multicolumn{1}{l|}{UDG} & 70.73 & 77.88 & \multicolumn{1}{c|}{74.30} & 85.95 & 81.79 & 78.54 & \multicolumn{1}{c|}{82.09} & 68.11 \\
\multicolumn{1}{l|}{MixOE} & 80.23 & 85.01 & \multicolumn{1}{c|}{82.62} & 90.64 & 86.80 & 87.36 & \multicolumn{1}{c|}{88.27} & 85.71 \\ \midrule
\multicolumn{9}{l}{\textit{\textbf{Training w/ Synthesized Outlier}}} \\ \midrule
\multicolumn{1}{l|}{VOS} & 79.68 & 85.35 & \multicolumn{1}{c|}{82.51} & 92.77 & 90.95 & 89.28 & \multicolumn{1}{c|}{91.00} & 86.23 \\
\multicolumn{1}{l|}{CIDER} & 76.04 & 85.13 & \multicolumn{1}{c|}{80.58} & 90.69 & 92.38 & 88.92 & \multicolumn{1}{c|}{90.66} & - \\
\multicolumn{1}{l|}{NPOS} & 74.29 & 84.50 & \multicolumn{1}{c|}{79.40} & 94.81 & \textbf{96.97} & 91.69 & \multicolumn{1}{c|}{94.49} & - \\
\multicolumn{1}{l|}{DreamOOD} & 75.89 & 85.36 & \multicolumn{1}{c|}{80.62} & 92.58 & 92.64 & 91.65 & \multicolumn{1}{c|}{92.29} & 85.79 \\  \midrule
\rowcolor{cyan!15} \multicolumn{1}{l|}{SONA (Ours)} & \textbf{87.01}$\pm$0.12 & \textbf{89.76}$\pm$0.24 & \multicolumn{1}{c|}{\textbf{88.38}$\pm$0.18} & \textbf{95.93}$\pm$0.17 & 95.41$\pm$0.19 & \textbf{96.28}$\pm$0.24 & \multicolumn{1}{c|}{\textbf{95.87}$\pm$0.20} & 86.64$\pm$0.25 \\ \bottomrule
\end{tabular}
\caption{AUROC (\%) comparison of various OOD detection methods on ImageNet-200 as the ID dataset. This table presents the mean and standard deviations, averaged over five trials for each method.}
\vspace{-0.15in}
\label{tab:main1} 
\end{table*}

%% file: 4_experiments.tex
\section{Experiments} \label{sec:4}

In this section, we evaluate the OOD detection performance of our SONA framework. We first describe our experimental setups (Section~\ref{sec:4.1}), then showcase novel outlier examples and impressive main results (Section~\ref{sec:4.2}). Finally, we present various analyses proving the robustness of our framework (Section~\ref{sec:4.3}).

\subsection{Experimental Setup} \label{sec:4.1}

\subsubsection{Datasets.} We mainly evaluate our framework on ImageNet-200 \cite{Jingyang2023OpenOOD} as ID, a subset of 200 categories from ImageNet-1k \cite{Deng2009CVPR}. Our evaluation covers both far-OOD and challenging near-OOD scenarios. For far-OOD detection, we employ widely-used datasets such as iNaturalist, Texture, and OpenImage-O. For near-OOD detection and SSB-hard and NINCO for near-OOD detection., which have no class overlap but show close semantic similarity with ImageNet-1K.

\begin{figure}[ht]
  \centering
  \includegraphics[width=\linewidth]{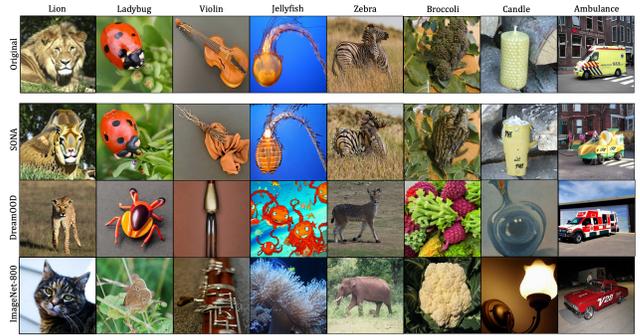} 
    \caption{\textbf{Comparison of original and synthesized outlier images}. SONA resemble ID mainly in nuisances and clearly represent semantically discrepant information, while others significantly deviate from the ID.}
    \vspace{-0.1in}
  \label{fig:gen}
\end{figure}

\subsubsection{Implementation details.} 
Our implementation is based on Stable Diffusion v2-base model \footnote{\label{ft:stable_diffusion}See \url{https://huggingface.co/stabilityai/stable-diffusion-2-base} for more details}, with sampling hyperparameters consistent with \cite{Brack2024SEGA}. For selecting $\mathbf{c}_\texttt{OOD}$, we utilize labels from the remaining 800 classes of ImageNet-1k (disjoint classes from ImageNet-200) for the main table. The number of generated samples is equal to the number of entries in the ID training dataset. For the OOD detector training, we employ ResNet-18 \cite{He2016ResNet} as the network architecture. Additional implementation details can be found in the Appendix.

\subsubsection{Evaluation metric.} We primarily assess our method through the Area Under the Receiver Operating Characteristic Curve (AUROC) and accuracy on ID data. To ensure reliability, all reported results are averaged across 5 random seeds for both outlier generation and detector training.

\subsection{Main Results} \label{sec:4.2}
Table \ref{tab:main1} presents a comprehensive comparison of OOD detection performance on ImageNet-200, highlighting the superiority of our framework. Our proposed method consistently outperforms in both near and far-OOD scenarios, affirming its effectiveness. It surpasses all benchmarks in the near-OOD setting, achieves top results in two out of three far-OOD datasets, and attains the second-best performance in the remaining one. Remarkably, our method achieves a state-of-the-art (SOTA) AUROC of 88.4\% on near-OOD detection, showing exceptional performance in this challenging scenario. This score represents a notable improvement, outperforming other synthesis-based methods by 6\%. Furthermore, while other baseline methods experience a sharp decline of 10\% in near-OOD compared to far-OOD, our approach significantly reduces this performance gap. Another noteworthy point is we even surpass real-world outlier based methods, without requiring any additional datasets. This indicates that our methodology is also cost-effectively feasible for real-world applications. Another superior performance of SONA across a broader range of settings, including full-spectrum (\emph{e.g.,} covariate-shifted ID) and fine-grained benchmarks, can be found in the Appendix.

We further compare our generated images visually with another diffusion-based outlier synthesis method, Dream-OOD, as well as with real-world samples from ImageNet-800, which also share similar representations with the ID. As shown in Figure~\ref{fig:gen}, Dream-OOD and ImageNet-800 display distinct information from the ID in terms of both semantics and nuisances. On the other hand, our generated samples closely resemble the ID primarily in nuisances, while at the same time, they successfully exhibit clear semantic discrepancies. Therefore, SONA effectively assists detectors in capturing even slight semantic differences with OOD by providing an intuitive understanding of the visual characteristics of images.

\begin{figure*}[t]
    \centering
    \begin{minipage}{0.48\textwidth}
        \centering
        \includegraphics[width=\linewidth]{aaai_figures/4_exp_timestep.pdf}
        \vspace{-0.3in}
        \caption{Performance comparison of SONA guidance and global guidance across $\tilde{T}$ on ImageNet200 (ID). (a) AUROC (\%) analysis. (b) LPIPS score analysis.}
        \label{exp_timesetp}
    \end{minipage}\hfill
    \begin{minipage}{0.48\textwidth}
        \vspace{-5mm} 
        \centering
        \includegraphics[width=\linewidth]{aaai_figures/4_hyper_plot.pdf}
        \vspace{-0.3in}
        \caption{(a) Ablation study on region masking threshold $\lambda$ on ImageNet200 (ID) (b) Ablation study of SONA guidance scale $s$.}
        \label{fig:4_hyper_plot}
    \end{minipage}
\end{figure*}
\vspace{-0.05in}
\subsection{Analysis} \label{sec:4.3}
\vspace{-0.05in}
\subsubsection{SONA remains competitive regardless of $\tilde{T}$.}
SONA guidance effectively mitigates the sensitivity of $\tilde{T}$, as shown in Figure~\ref{exp_timesetp}. Global guidance, which uniformly applies guidance to entire regions, shows performance variations depending on the $\tilde{T}$, with the best results at a fixed $\tilde{T}=25$. In contrast, SONA guidance demonstrates robust results across both early and later $\tilde{T}$ values, achieving the best score when randomly choosing $\tilde{T} \sim \mathcal{U}(1, 50)$ for each sample, exhibiting an ensemble effect. In addition, we evaluate the similarity between the ID and SONA samples at each timestep using the LPIPS \cite{Zhang2018CVPR} score. As timesteps increase, the global guidance's LPIPS score continues to rise, indicating growing dissimilarity. In contrast, SONA demonstrates minimal LPIPS score variations, effectively starting to remove original semantics in early $\tilde{T}$ and preserving nuisances until in later $\tilde{T}$. This property eliminates the need for meticulous $\tilde{T}$ tuning, reducing the dependency on $\tilde{T}$ and making our approach more robust compared to methods heavily relying on optimal hyperparameters.

\subsubsection{SONA remains competitive regardless of $\mathbf{c}_\texttt{OOD}$.}
SONA presents remarkable consistent results across different $\mathbf{c}_\texttt{OOD}$ selection methods (Table \ref{tab:ood_prompt}), while DreamOOD is highly sensitive to the selection of OOD prompts. This robustness is due to SONA's unique approach of using ID images directly, unlike general diffusion methods starting from random noise. By leveraging intrinsic ID characteristics, SONA effectively captures subtle variations crucial for Near-OOD detection.

\begin{table}[t]
\centering
\small
\renewcommand{\arraystretch}{0.9}
\begin{tabular}{@{}lcc@{}}
\toprule
\multirow{2}{*}{\centering \textbf{OOD prompt}} & \multicolumn{2}{c}{\textbf{AUROC (\%)}} \\ \cmidrule{2-3}
 & Near-OOD       & Far-OOD        \\ \midrule
LAION                      & 87.33          & 94.65          \\
ImageNet-800 - close       & 88.38          & 94.60          \\
ImageNet-800 - far         & 87.31          & 94.47          \\
\rowcolor{cyan!15}ImageNet-800 - rand (Ours) & \textbf{88.38} & \textbf{95.87} \\ \bottomrule
\end{tabular}
\caption{Comparison of SONA with different OOD prompt selection.}
\label{tab:ood_prompt}
\vspace{-0.2in}
\end{table}

\subsubsection{Ablation of the components in $\Delta_\text{SONA}$.}
The ablation study in Table~\ref{tab:masking} reveals the progressive impact of SONA's components. $\Delta_\text{ID}$ excludes ID semantics from early timesteps, showing slight performance improvement. $\Delta_\text{OOD}$ with nuisance filtering facilitates semantic corruption, however, the absence of filtering results in performance decline, especially for Near-OOD settings. The full $\Delta_\text{SONA}$ balances both semantic corruption and nuisance remaining, resulting in consistent and precise outlier generation.

\subsubsection{Ablation of the components in $\mathcal{L}$} 
We conducted an ablation study to examine the impact of loss function combinations (Table~\ref{tab:ablation_loss}). The exposure of SONA outliers to the classifier through $\mathcal{L}_{\texttt{CE}}+\mathcal{L}_{\texttt{OE}}$ demonstrated significant performance improvements across near-to-far scenarios compared to $\mathcal{L}_{\texttt{CE}}$, strongly validating our approach. Furthermore, the addition of $\mathcal{L}_{\texttt{MI}}$ further enhanced the classifier learning, reinforcing the robustness of our method.

\subsubsection{Ablation on the hyperparameters of SONA guidance.}
To investigate the impact of our SONA guidance, we conduct an ablation study of the region masking threshold $\lambda$ and the guidance scale $s$ (Figure~\ref{fig:4_hyper_plot}). We evaluate across a range of values for $\lambda \in \{0.1, 0.15, 0.2, 0.25\}$ and $s \in \{5, 10, 15, 20\}$. The results demonstrate that our framework is robust and not sensitive to variations in both hyperparameters, with only minor fluctuations in the AUROC scores. This robustness makes SONA more practical and reliable for real-world applications.

\begin{table}
\centering
\renewcommand{\arraystretch}{0.85}
\setlength{\tabcolsep}{2.5pt}
\small
\begin{tabular}{@{}cccccc@{}}
\toprule
\multicolumn{4}{c}{\textbf{$\Delta_\texttt{SONA}$ components.}} & \multicolumn{2}{c}{\textbf{AUROC (\%)}} \\ 
\cmidrule(lr){1-4} \cmidrule(lr){5-6}
$\Delta_\text{ID}$& \makecell{$\Delta_\text{OOD}$ \\ (\emph{w. N filtering})} & \makecell{$\Delta_\text{OOD}$ \\ (\emph{w/o. N filtering})} &$\Delta_\text{N}$&Near-OOD&Far-OOD\\ \midrule
\rowcolor{gray!20} \multicolumn{4}{c}{\textit{Global guidance}} & 82.85&91.19\\
\textbf{\checkmark}&&&&83.74&93.42\\
\textbf{\checkmark}&\textbf{\checkmark}&&&84.56&94.21\\
\textbf{\checkmark}&&\textbf{\checkmark}&&82.57&93.89\\
\rowcolor{cyan!15} \textbf{\checkmark}&\textbf{\checkmark}&&\textbf{\checkmark}&\textbf{88.38}&\textbf{95.87}\\ \bottomrule
\end{tabular}
\caption{Ablation study on $\Delta_\texttt{SONA}$ components.}
\label{tab:masking}
\end{table}


\begin{table}[t]
\centering
\small
\renewcommand{\arraystretch}{0.9}
\begin{tabular}{@{}lcc@{}}
\toprule
\multirow{2}{*}{\centering \textbf{Loss function}} & \multicolumn{2}{c}{\textbf{AUROC (\%)}} \\ \cmidrule{2-3}
& Near-OOD & Far-OOD \\ \midrule
$\mathcal{L}_{\texttt{CE}}$ & 82.5 & 90.86 \\
$\mathcal{L}_{\texttt{CE}} + \mathcal{L}_{\texttt{OE}}$ & 87.19 & 94.52 \\
\rowcolor{cyan!15} $\mathcal{L}_{\texttt{CE}} + \mathcal{L}_{\texttt{OE}} + \mathcal{L}_{\texttt{MI}}$ & \textbf{88.38} & \textbf{95.87} \\ \bottomrule
\end{tabular}
\caption{Ablation study on $\mathcal{L}$ components.}
\vspace{-0.2in}
\label{tab:ablation_loss}
\end{table}

%% file: 5_relatedwork.tex
\vspace{-0.05in}
\section{Related Work}
\subsection{OOD Detection with Auxiliary Outliers} \label{sec:6.1}
Recent strategies aim to construct robust OOD detectors by regularizing the classifier with real-world outliers in training OE \cite{Hendrycks2019Deep}. \cite{yu2019unsupervised} trains two randomly initialized classifiers and minimizes their discrepancy on the outlier. \cite{Jin2021Seman} collects semantically coherent outliers through clustering. \cite{Jin2023Mixture} synthesize fine-grained outlier by applying mixup \cite{Zhang2018MixUP}. However, these methods require explicit outlier gathering, which may be costly. An alternative approach is to synthesize outliers on pixel or latent space. \cite{Du2022VOS} approximates the ID latent distribution as a mixture of class-conditional Gaussians and sample outliers deviating from this mixture. \cite{Tao2023NPOS} identifies boundary ID samples in the CLIP \cite{Radford21} space and regards distant features as outliers. \cite{mirzaei2022fake} obtain outliers by early stopping diffusion model scratch training. \cite{du2024dream} generates pixel-space outliers using diffusion models with CLIP space distance information.
\vspace{-0.05in}
\subsection{Semantic Guidance for Diffusion Model} \label{sec:6.4}
Text-guided diffusion models allow for controlling semantic content through textual prompts. One approach to enhancing fine-grained controllability is inpainting \cite{nichol2021glide, couairon2022diffedit}, which uses pre-defined or learnable masks to modify the semantic regions of an image. Another line of research has focused on developing more semantically grounded approaches, leveraging the semantics encoded in the cross-attention maps \cite{hertz2022prompt} of the diffusion model or directly manipulating noise estimates \cite{Brack2024SEGA}. Instead of using semantic control for image editing, we aim to leverage it for outlier generation, intentionally inducing imperfect semantic control to produce outlier samples that slightly deviate from the original meaning.

%% file: 6_conclusion.tex

\vspace{-0.15in}
\section{Conclusion}
In this paper, we introduce \emph{Semantic Outlier generation via Nuisance Awareness (SONA)}, a novel and effective outlier synthesizing framework for OOD detection. Our key idea is to leverage the informative pixel-space ID images for outlier generation by directly incorporating them into diffusion models. To this end, we propose \emph{SONA guidance} that enables the generation of diverse outliers that closely resemble the ID in nuisance regions while representing semantically distinct information. By training OOD detectors with SONA samples, we successfully capture subtle distinctions between ID and OOD, leading to improved OOD detection performance.


\section*{Acknowledgements}
This work is fully supported by LG AI Research.

%% file: 7_supplement.tex
\appendix
\onecolumn

\begin{center}
    \LARGE\textbf{Supplementary Material}
\end{center}
\begin{center}
    \Large\textbf{Diffusion-based Semantic Outlier Generation via
Nuisance Awareness \\ for Out-of-Distribution Detection}
\end{center}

\section{Backgrounds}
\label{sup_sec:diff}
\subsection{Diffusion Models}
A diffusion model is a generative model that gradually adds noise to an input signal $\mathbf{x}=\mathbf{x}_0$ until it is fully destroyed into random noise $\mathbf{x}_T$, and then denoises it through multiple steps to generate an output signal $\tilde{\mathbf{x}}_0$ with a probability distribution similar to that of the input. A diffusion process is defined as a Gaussian process with a Markov chain:

\begin{equation}
    \mathbf{x}_{t} = \sqrt{1-\beta_{t}}\mathbf{x}_{t-1} + \sqrt{\beta_{t}}\mathbf{z}_{t},  t=1,...,T
\end{equation}

where $\beta_{1},...,\beta_{T}$ is a fixed variance scheduler which means the quantity of noise for each step $t$ and $\mathbf{z}_{t} \sim \mathcal{N}(0,I)$.
It can be rewritten as,

\begin{equation}
    q(\mathbf{x}_{t} | \mathbf{x}_{t-1}) = \mathcal{N}(\mathbf{x}_{t};\sqrt{1-\beta_{t}}\mathbf{x}_{t-1},\beta_{t}I)
\end{equation}
\begin{equation}
    q(\mathbf{x}_{t} | \mathbf{x}_{0}) = \mathcal{N}(\mathbf{x}_{t};\sqrt{\bar{\alpha}_{t}}\mathbf{x}_{t-1},(1-\bar{\alpha}_{t})I)
\end{equation}

where $\alpha_{t} \coloneqq 1-\beta_{t}$ and $\bar{\alpha}_{t} \coloneqq \prod_{s=1}^{t}\alpha_{s}$.

To recover the input signal, we need to learn reverse process, which requires estimating the noise prediction function $\epsilon_{\theta}(\mathbf{x}_{t}); t=1,...,T$. The parameter $\theta$ is optimized by minimizing follows:

\begin{equation}
    \mathcal{L}(\theta)=\mathbb{E}_{\epsilon ,\mathbf{x},t}[\left\| \epsilon_{\theta}(\mathbf{x}_{t})-\epsilon \right\|^{2}_{2}]
\end{equation}

in which $\epsilon \sim \mathcal{N}(0,I)$. This objective performs denoising score matching over multiple noise scales by $t$. Leveraging predicted noise $\epsilon_{\theta}$, we can sample $\mathbf{x}_{t-1} \sim p(\mathbf{x}_{t-1}|\mathbf{x}_{t})$. The most widely adopted sampling method is Denoising Diffusion Probabilistic Models (DDPM) \cite{ho2020denoising} sampler: 

\begin{equation}
 \mathbf{x}_{t-1}=\frac{1}{\sqrt{\alpha}_{t}}(\mathbf{x}_{t}-\frac{1-\alpha_{t}}{\sqrt{1-\bar{\alpha}_{t}}})\epsilon _{\theta}(\mathbf{x}_t)+\sigma_{t}\mathbf{z}
\end{equation}

\subsection{Classifier-Free Diffusion Guidance}
Classifier-free Diffusion Guidance (CFG) \cite{ho2022classifier} is a simple yet effective conditional diffusion model that avoids the need for a separate classifier. They obtain a combination of a conditional model parameterized with $\epsilon_{\theta}(\mathbf{x}_{t}, \mathbf{c})$ and an unconditional model parameterized with $\epsilon_{\theta}(\mathbf{x}_{t})=\epsilon_{\theta}(\mathbf{x}_{t}, \mathbf{c} = \varnothing)$, which gives a null token to guidance $\mathbf{c}$ in a single network. During training, it randomly drops the condition with an unconditional probability $p_{\texttt{uncond}}$. The training process is described in Algorithm~\ref{algo:train}.

\begin{algorithm}
\footnotesize
    \caption{Classifier-Free Diffusion Guidance Training} 
    \label{algo:train}
    \begin{algorithmic}
        \Require $p_{\texttt{uncond}}$: probability of unconditional training
        \Require $\mathbf{c}$: conditional guidance signal 
        \Repeat
            \State $(\mathbf{x}, \mathbf{c}) \sim p(\mathbf{x}, \mathbf{c})$
            \State $\mathbf{c} \to \emptyset $ with probability $p_{\texttt{uncond}}$
            \State $\lambda \sim p(\lambda)$
            \State $\epsilon \sim \mathcal{N}(0,I)$
            \State $z_{\lambda } =\alpha _{\lambda}\mathbf{x} + \sigma_{\lambda }\epsilon$
            \State Take gradient step on $\nabla_{\theta }\left\| \epsilon_{\theta }(\mathbf{z}_{\lambda }, \mathbf{c})- \epsilon \right\|^{2}$
        \Until{converged}
    \end{algorithmic} 
\end{algorithm}


\section{Dataset Details}

\subsection{Near-OOD Detection}

For Near-OOD detection, we utilize two specific OOD datasets that have been carefully chosen to challenge the robustness of our model: SSB-hard and NINCO. This setting is adopted directly from the OpenOOD v1.5 \cite{Jingyang2023OpenOOD} framework, ensuring consistency with established benchmarks.

\begin{itemize}
    \item \textbf{SSB-hard \cite{vaze2022open} (Semantic Shift Benchmark - Hard)}:
    SSB-hard is a subset derived from the Semantic Shift Benchmark (SSB), designed to evaluate model performance under subtle semantic shifts. It contains images closely similar to ImageNet-1K, making it particularly challenging. The dataset comprises 980 classes from ImageNet-21K, sharing significant visual and semantic characteristics with ImageNet-1K.

    \item \textbf{NINCO \cite{bitterwolf2023or} (No ImageNet-1K Class Objects)}:
        NINCO is a new dataset with 64 unique classes designed for robust OOD detection evaluation. These classes do not overlap with ImageNet-1K but are visually similar, testing the model's ability to generalize beyond training data. It helps assess the model's performance in identifying visually similar OOD samples without relying on class mismatch.

\end{itemize}

\subsection{Far-OOD Detection}

For far-OOD detection, we leverage OOD test datasets that are commonly used in prior OOD detection research. These datasets provide a stark contrast to the ID data, thereby testing the model's ability to handle significantly different OOD scenarios.

\begin{itemize}
    \item \textbf{iNaturalist \cite{van2018inaturalist}}:
    iNaturalist is a large-scale, diverse dataset containing images of various species of plants, animals, and other natural entities. It is used to validate the model's performance in detecting OOD samples significantly different from ImageNet-1K, ensuring the model can handle a wide range of natural variations.

    \item \textbf{Textures \cite{cimpoi2014describing}}:
    The Textures dataset contains images representing a wide variety of real-world patterns and surface textures, differing from the object-centric images in ImageNet-1K. It tests the model's ability to recognize OOD samples based on textural differences, crucial for robust OOD detection.

    \item \textbf{OpenImage-O \cite{Wang2022VIM}}:
    OpenImage-O, a subset of the OpenImages dataset curated for OOD detection research, includes diverse images not in ImageNet-1K, offering broad OOD scenarios. It assesses the model's capability to detect OOD instances in an open-world setting with a greater variety of objects and scenes.

\end{itemize}

\section{Implementation Details of SONA Framework} \label{sppl:impl_detail}

\paragraph{Implementation details.} Our implementation for SONA is based on the Stable Diffusion v2-base model. For more information, please refer to Footnote~\ref{ft:stable_diffusion}. For our best results, we set the total number of diffusion steps $T$ to 50, and at that point randomly selected a stop timestep $\Tilde{T}$ from the range between 1 and 50. For the OOD text condition, $\textbf{c}_{\texttt{OOD}}$, we use text prompts based on ImageNet-800 words, which do not overlap with ImageNet-200. We set the guidance scale $s$ to 10 and the guidance threshold $\lambda$ to 0.2 for all datasets. We follow the additional hyperparameters for image sampling provided by \cite{Brack2024SEGA}. Since we use the ID training dataset directly for SONA generation, the number of SONA is equal to the number of entries in the ID training dataset. While synthesizing our SONA samples, we use 8 A40 gpus. Implementation of SONA is summarized as following algorithm \ref{algo:sona}:

For our main results on ImageNet-200, following \cite{Jingyang2023OpenOOD}, we train a ResNet-18 \cite{He2016ResNet} classifier from scratch for 100 epochs with a batch size of 256 and resized images of 224. The optimizer used is SGD with a momentum of 0.9. We employ a learning rate of 0.1 with a cosine annealing decay schedule. For the fine-grained setting, we also train a ResNet-18  classifier from scratch for 100 epochs with a batch size of 32 and resized images of 448. The optimizer used is SGD with a momentum of 0.9. We employ a learning rate of 0.1 with a cosine annealing decay schedule. 

\paragraph{Baselines.} We compare our framework against existing methods divided into three categories: \textit{Post-hoc} detection methods that do not require further training, training with real-world outlier methods that require explicit outliers, and outlier synthesis-based methods. For the \textit{post-hoc} baselines, we compare against OpenMax , MSP, ODIN, EBO, OpenGAN , ReAct, KNN , and DICE. For the methods that require explicit outlier, our comparison includes OE, MCD, UDG, and MixOE where all methods use remaining 800 categories of the ImageNet-1k dataset as additional outliers. Finally, we compare against various OOD synthesis methods, including VOS, CIDER , NPOS and DreamOOD.

\section{Computational Cost}
We summarize the computational cost of \textbf{SONA}. For outlier synthesis, generating a total of 200K images takes approximately 8.1 hours. This is significantly less time compared to other outlier-synthesis methods, as our approach does not generate outliers from random noise with full DDIM steps, but rather starts from the original image and stops early. Additionally, SONA demonstrates superior computational efficiency compared to methods that calculate perceptual distances between input images and their diffusion-generated counterparts \cite{liu2023unsupervised,graham2023denoising, gao2023diffguard}. During inference, SONA is substantially faster than these methods, which require computationally intensive diffusion reverse processes with multiple backward steps (e.g., 80x faster vs. DiffGuard \cite{gao2023diffguard}). This efficiency is particularly advantageous for continuously tracking and evaluating multiple datasets. Furthermore, SONA can operate effectively in on-device deployments where large diffusion models are impractical.

For classifier training, our approach requires approximately 5 hours when training from scratch with the generated outliers and an additional OE loss term. In comparison, post-hoc methods take approximately 4 hours to train a multi-class classification model on the ImageNet-200 training dataset. For ImageNet-1K, we leverage a pre-trained classifier and simply fine-tune it for SONA within a few epochs, taking only about 1 hour on 8 A100 GPUs while achieving superior performance. This efficient fine-tuning process effectively addresses concerns about classifier training costs. Our experiments were conducted using NVIDIA GeForce RTX 3090Ti GPUs.

\section{Full Algorithm of SONA}

\begin{algorithm}[htbp]
\caption{Semantic Outlier generation via Nuisance Awareness (SONA)}
\label{algo:sona}
\begin{algorithmic}[1]
\Input In-distribution training data $\mathcal{D}_{\texttt{ID}} = {(\mathbf{x}^{(i)}, y^{(i)})}_{i=1}^n$
\Output SONA $\mathcal{D}_{\texttt{OOD}}=\{\mathbf{x}_{\texttt{OOD}}^{(i)}\}_{i=1}^n$.
\Require Latent Diffusion Model $LDM$ with initial weights $\theta$, Diffusion timestep $T$
\Require ID prompt embedding $\mathbf{c}_\texttt{ID}$, OOD prompt embedding $\mathbf{c}_\texttt{OOD}$, threshold $\lambda$
\State $\Tilde{T} \sim \mathcal{U}(0, T)$
\State $\mathbf{z} \gets LDM.encode(\mathbf{x})$ 
\State $\mathbf{z}_{\Tilde{T}} \gets LDM.add-noise(\mathbf{z}, \Tilde{T})$ \Comment{Add noise at timestep $\Tilde{T}$}
\For{$t \gets \Tilde{T}, \ldots, 1$}
\State $\psi_{\mathbf{c}_\texttt{ID}}, \psi_{\mathbf{c}_\texttt{OOD}} \gets LDM.forward(\mathbf{z}_t, \mathbf{c}_\emptyset, \mathbf{c}_\texttt{ID}, \mathbf{c}_\texttt{OOD})$
\State $M_\texttt{S\_ID}, M_\texttt{S\_OOD}, M_\texttt{N\_ID} \gets \texttt{GetMasks}(\mathbf{z}_t, \mathbf{c}_\texttt{ID}, \mathbf{c}_\texttt{OOD}, \eta_\lambda)$
\State $\Delta_\texttt{ID} \gets -M_\texttt{S\_ID} \odot \psi_{\mathbf{c}_\texttt{ID}}$ \Comment{Remove ID semantic}
\State $\Delta_\texttt{N} \gets M_\texttt{N\_ID} \odot \psi_{\mathbf{c}_\texttt{ID}}$ \Comment{Preserve ID nuisance}
\State $\Delta_{\texttt{OOD}} \gets M_\texttt{S\_OOD} \odot (1-M_\texttt{N\_ID}) \odot \psi_{\mathbf{c}_\texttt{OOD}}$ \Comment{Add OOD semantic}
\State $\Delta \gets s \cdot (\Delta_{\texttt{ID}} + \Delta_{\texttt{OOD}} + \Delta_\texttt{N})$
\State $\mathbf{z}_{t-1} \gets LDM.denoise(\mathbf{z}_t, \Delta)$ \Comment{Denoise with SONA guidance}
\EndFor
\State $\mathbf{x}_{\texttt{OOD}} \gets LDM.decode(\mathbf{z}_0)$
\State \textbf{return} $\mathbf{x}_{\texttt{OOD}}$
\end{algorithmic}
\end{algorithm}

\section{Additional OOD Detection Empirical Results}
\label{supp:secb}

\subsection{Other OOD Detection Benchmarks: CIFAR-100}
We extend the evaluation of our framework to another benchmark, the CIFAR-100 ~\cite{krizhevsky2009learning} dataset as ID, to demonstrate the further application of our framework. 

\paragraph{Dataset Details.} CIFAR-100 contains 100 different classes, with each class containing 500 images. We use TIN as the near-OOD dataset and MNIST, SVHN, Texture, and Place365 as far-OOD datasets. For the SONA outlier generation phase, we use CIFAR-10 as the OOD label set and generate the same number of samples as in the ID training dataset. We validated using the same implementation details as described in Appendix~\ref{sppl:impl_detail}.

\paragraph{Results.} Table \ref{tab:main2_detail} presents a comprehensive comparison of OOD detection performance on CIFAR-100, further validating the effectiveness of our framework. Our proposed method consistently demonstrates superior performance across both near and far-OOD scenarios. In the challenging near-OOD setting, our approach outperforms all baselines. For far-OOD detection, our method achieves top results in the majority of the datasets evaluated (SVHN, Texture and Place365). It's particularly noteworthy that our method outperforms approaches that rely on real-world outlier datasets, despite not requiring any additional data beyond the ID training set and synthetically generated OOD samples. This highlights the cost-effectiveness and practical applicability of our methodology in real-world scenarios. These results on CIFAR-100 strongly corroborate our findings from the ImageNet-200 evaluation, underscoring the robustness of our framework across different dataset scales.

\begin{table}[htbp]
\small
\centering
\setlength{\tabcolsep}{10pt}
\renewcommand{\arraystretch}{1.0}
\begin{tabular}{@{}lcccccc@{}}
\toprule
\multicolumn{1}{c}{\multirow{2}{*}{Method}} & Near-OOD & \multicolumn{4}{c}{Far-OOD} & \multirow{2}{*}{ID ACC} \\ \cmidrule(lr){2-6}
\multicolumn{1}{c}{} & TIN & MNIST & SVHN & Texture & Place365 &  \\ \midrule
\multicolumn{7}{l}{\textit{\textbf{Post-hoc}}} \\ \midrule
\multicolumn{1}{l|}{OpenMax} & \multicolumn{1}{c|}{78.44} & 76.01 & 82.07 & 80.56 & \multicolumn{1}{c|}{79.29} & 77.25 \\
\multicolumn{1}{l|}{MSP} & \multicolumn{1}{c|}{82.07} & 76.08 & 78.42 & 77.32 & \multicolumn{1}{c|}{79.22} & 77.25 \\
\multicolumn{1}{l|}{TempScale} & \multicolumn{1}{c|}{82.79} & 77.27 & 79.79 & 78.11 & \multicolumn{1}{c|}{79.8} & 77.25 \\
\multicolumn{1}{l|}{ODIN} & \multicolumn{1}{c|}{81.63} & 83.79 & 74.54 & 79.33 & \multicolumn{1}{c|}{79.45} & 77.25 \\
\multicolumn{1}{l|}{MDS} & \multicolumn{1}{c|}{61.5} & 67.47 & 70.68 & 76.26 & \multicolumn{1}{c|}{63.15} & 77.25 \\
\multicolumn{1}{l|}{MDSEns} & \multicolumn{1}{c|}{48.78} & 98.21 & 53.76 & 69.75 & \multicolumn{1}{c|}{42.27} & 77.25 \\
\multicolumn{1}{l|}{RMDS} & \multicolumn{1}{c|}{82.55} & 79.74 & 84.89 & 83.65 & \multicolumn{1}{c|}{83.4} & 77.25 \\
\multicolumn{1}{l|}{Gram} & \multicolumn{1}{c|}{53.91} & 80.71 & 95.55 & 70.79 & \multicolumn{1}{c|}{46.38} & 77.25 \\
\multicolumn{1}{l|}{EBO} & \multicolumn{1}{c|}{82.76} & 79.18 & 82.03 & 78.35 & \multicolumn{1}{c|}{79.52} & 77.25 \\
\multicolumn{1}{l|}{OpenGAN} & \multicolumn{1}{c|}{68.74} & 68.14 & 68.4 & 65.84 & \multicolumn{1}{c|}{69.13} & 77.25 \\
\multicolumn{1}{l|}{GradNorm} & \multicolumn{1}{c|}{69.95} & 65.35 & 76.95 & 64.58 & \multicolumn{1}{c|}{69.69} & 77.25 \\
\multicolumn{1}{l|}{ReAct} & \multicolumn{1}{c|}{82.88} & 78.37 & 83.01 & 80.15 & \multicolumn{1}{c|}{80.03} & 77.25 \\
\multicolumn{1}{l|}{MLS} & \multicolumn{1}{c|}{82.9} & 78.91 & 81.65 & 78.39 & \multicolumn{1}{c|}{79.75} & 77.25 \\
\multicolumn{1}{l|}{KLM} & \multicolumn{1}{c|}{79.22} & 74.15 & 79.34 & 75.77 & \multicolumn{1}{c|}{75.7} & 77.25 \\
\multicolumn{1}{l|}{VIM} & \multicolumn{1}{c|}{77.76} & 81.89 & 83.14 & 85.91 & \multicolumn{1}{c|}{75.85} & 77.25 \\
\multicolumn{1}{l|}{KNN} & \multicolumn{1}{c|}{83.34} & 82.36 & 84.15 & 83.66 & \multicolumn{1}{c|}{79.43} & 77.25 \\
\multicolumn{1}{l|}{DICE} & \multicolumn{1}{c|}{80.72} & 79.86 & 84.22 & 77.63 & \multicolumn{1}{c|}{78.33} & 77.25 \\ \midrule
\multicolumn{7}{l}{\textit{\textbf{Training w/ Real-World Outlier}}} \\ \midrule
\multicolumn{1}{l|}{OE} & \multicolumn{1}{c|}{79.62} & 80.68 & 84.37 & 82.18 & \multicolumn{1}{c|}{78.39} & 76.84 \\
\multicolumn{1}{l|}{MCD} & \multicolumn{1}{c|}{78.75} & 68.25 & 75.92 & 77.07 & \multicolumn{1}{c|}{77.65} & 75.83 \\
\multicolumn{1}{l|}{UDG} & 80.9 & \textbf{83.88} & 79.8 & 75.57 & \multicolumn{1}{c|}{79.11} & 71.54 \\
\multicolumn{1}{l|}{MixOE} & \multicolumn{1}{c|}{83.73} & 76.06 & 72.28 & 77.34 & \multicolumn{1}{c|}{79.92} & 75.13 \\ \midrule
\multicolumn{7}{l}{\textit{\textbf{Training w/ synthesized Outlier}}} \\ \midrule
\multicolumn{1}{l|}{VOS} & \multicolumn{1}{c|}{82.73} & 82.29 & 84.23 & 78.41 & \multicolumn{1}{c|}{80.34} & 77.2 \\
\multicolumn{1}{l|}{CIDER} & \multicolumn{1}{c|}{78.65} & 68.14 & 97.17 & 82.21 & \multicolumn{1}{c|}{74.43} & / \\
\multicolumn{1}{l|}{NPOS} & \multicolumn{1}{c|}{81.32} & 73.26 & 92.43 & 85.55 & \multicolumn{1}{c|}{77.92} & / \\
\multicolumn{1}{l|}{DreamOOD} & \multicolumn{1}{c|}{80.56} &  73.65&  96.23&  90.34& \multicolumn{1}{c|}{87.43} &  74.22\\ \midrule
\rowcolor{cyan!15}\multicolumn{1}{l|}{SONA (Ours)} & \multicolumn{1}{c|}{\textbf{84.07}$\pm$0.17} & 80.78$\pm$0.28 & \textbf{99.38}$\pm$0.19 & \textbf{92.25}$\pm$0.45 & \multicolumn{1}{c|}{\textbf{96.7}$\pm$0.21} & 77.95$\pm$0.30 \\ \bottomrule
\end{tabular}
\caption{AUROC (\%) comparison of various OOD detection methods on CIFAR-100 as the ID dataset.}
\label{tab:main2_detail} 
\end{table}

\subsection{Other OOD Detection Benchmarks: ImageNet-1k}
We extend the evaluation of our framework to another large-scale benchmark, the ImageNet-1k ~\cite{krizhevsky2009learning} dataset as ID, to demonstrate the further application of our framework. 

\paragraph{Dataset Details.} 
ImageNet-1K contains 1,000 different classes, with approximately 1.2 million training images. It includes more high-resolution and diverse classes than ImageNet-200, making it suitable for large-scale evaluations. We use the same far-OOD and near-OOD datasets as in the ImageNet-200 experiments. For the SONA outlier generation phase, we adopted LAION dataset for OOD prompt selection that are semantically distinct from the original ImageNet classes, ensuring effective outlier generation.

\begin{table}[htbp]
\centering
\resizebox{0.9\columnwidth}{!}{
\small
\begin{tabular}{@{}lcccccccc@{}}
\toprule
\multicolumn{1}{c}{\multirow{2}{*}{Method}}  & \multicolumn{3}{c}{Near-OOD} & \multicolumn{4}{c}{Far-OOD}                  & \multirow{2}{*}{ID ACC} \\ \cmidrule(lr){2-8}
                        & SSB-hard   & NINCO  & Avg    & iNaturalist & Textures & OpenImage-O & Avg   &                         \\ \midrule
OpenMax                 & 71.37      & 78.17  & 74.77  & 92.05       & 88.10    & 87.62       & 89.26 & 76.18                   \\
MSP                     & 72.09      & 79.95  & 76.02  & 88.41       & 82.43    & 84.86        & 85.23 & 76.18                   \\
ODIN                    &  71.74     & 77.77  & 74.75  & 91.17       & 89.00     & 88.23       & 89.47 & 76.18                   \\
EBO                     & 72.08      & 79.70  & 75.89  & 90.63       & 88.70    & 89.06       & 89.47 & 76.18                   \\
VOS                     & 76.26      & 80.25  & 78.26  & 92.34       & 91.13    & 90.17       & 91.21 & 77.13                   \\
OE                      & 78.60      & 86.70  & 82.65  & 90.76       & 89.45    & 89.01       & 89.74  & 76.02                   \\
MixOE                   & 73.32      & 82.66  & 77.99  & 91.25       & 89.91    & 88.67       & 89.94  & 75.93                   \\
DreamOOD                & 72.07      & 81.98  & 77.03  & 93.18       & 92.53    &  91.78    &  92.50    &  75.43                  \\ \midrule
\rowcolor{cyan!15} SONA (Ours)             & \textbf{82.21}$\pm$0.15      & \textbf{88.43}$\pm$0.21  & \textbf{85.32}$\pm$0.18 & \textbf{95.56}$\pm$0.14       & \textbf{93.21}$\pm$0.28    & \textbf{95.69}$\pm$0.16       & \textbf{94.82}$\pm$0.19 & 76.94 $\pm$0.23                 \\ \bottomrule
\end{tabular}
}
\caption{AUROC (\%) comparison of various OOD detection methods on ImageNet-1k as the ID dataset.}
\label{table:in1k}
\end{table}

\paragraph{Results.} 
Table \ref{table:in1k} presents a comprehensive comparison of OOD detection performance on ImageNet-1K. SONA demonstrates significant improvements in AUROC across both near- and far-OOD settings compared to existing baselines. The effectiveness of our framework extends to large-scale evaluations, with SONA consistently generating high-quality corrupted outliers even when using prompts from the LAION dataset. These results validate our method's robustness across different OOD prompt sources and its scalability to comprehensive datasets like ImageNet-1K.

\subsection{Robustness OOD Detection Benchmarks: ImageNet-200 Full-spectrum}
To evaluate the robustness of our framework in full-spectrum OOD detection, we extend our validation to encompass covariate-shifted ID datasets.

\paragraph{Dataset Details.} We use ImageNet-200 as our base ID dataset, and incorporate ImageNet-C~\cite{hendrycks2019benchmarking}, ImageNet-R~\cite{hendrycks2021many}, and ImageNet-V2~\cite{recht2019imagenet} as test ID datasets exhibiting covariate shift. These datasets introduce various types of distributional shifts, including corruptions, rendering styles, and natural variations, while maintaining the same semantic information with the original ImageNet-200. This setup allows us to assess our framework in distinguishing between ID samples with covariate shift and true out-of-distribution samples. We validated using the same implementation details as described in Appendix~\ref{sppl:impl_detail}.

\paragraph{Results.} Table \ref{table:fs} presents the comprehensive results of our full-spectrum OOD detection evaluation. Our proposed method consistently demonstrated strong performance across all covariate-shifted scenarios, underscoring its adaptability to various types of distributional shifts. These findings highlight our framework's robustness in challenging environments and its potential for real-world applications where both semantic and non-semantic shifts are prevalent. The results align with our observations from the standard OOD detection setting, reinforcing the robustness and broad applicability of our approach across different evaluation settings.

\begin{table}[htbp]
\centering
\resizebox{0.9\columnwidth}{!}{
\small 
\begin{tabular}{@{}lcccccccc@{}}
\toprule
\multicolumn{1}{c}{\multirow{2}{*}{Method}}  & \multicolumn{3}{c}{Near-OOD} & \multicolumn{4}{c}{Far-OOD}                  & \multirow{2}{*}{ID ACC} \\ \cmidrule(lr){2-8}
                        & SSB-hard   & NINCO  & Avg    & iNaturalist & Textures & OpenImage-O & Avg   &                         \\ \midrule
OpenMax                 & 47.64      & 54.15  & 50.89  & \textbf{72.44}       & 69.12    & 62.31       & 67.96 & 43.92                   \\
MSP                     & 50.94      & 57.76  & 54.35  & 70.42       & 65.11    & 62.8        & 66.11 & 43.92                   \\
ODIN                    & 44.31      & 52.36  & 48.33  & 70.19       & 67.1     & 61.48       & 66.25 & 43.92                   \\
EBO                     & 47.56      & 53.45  & 50.51  & 65.91       & 68.03    & 59.83       & 64.59 & 43.92                   \\
VOS                     & 47.01      & 53.27  & 50.14  & 65.95       & 67.83    & 59.54       & 64.44 & 43.92                   \\
OE                      & 54.31      & 58.85  & 56.58  & 61.08       & 60.75    & 60.57       & 60.8  & 44.29                   \\
MixOE                   & 49.88      & 54.24  & 52.06  & 63.52       & 60.44    & 58.04       & 60.66 & 44.36                   \\
DreamOOD                & 49.32      & 53.22  & 51.27  & 66.71       & 64.13    & 66.91       & 65.91 & 42.91                   \\ \midrule
\rowcolor{cyan!15} SONA (Ours)             & \textbf{56.26}$\pm$0.26      & \textbf{60.17}$\pm$0.13  & \textbf{58.22}$\pm$0.20 & 70.42$\pm$0.33       & \textbf{69.24}$\pm$0.24    & \textbf{72.43}$\pm$0.18       & \textbf{70.69}$\pm$0.25 & 44.52$\pm$0.16                  \\ \bottomrule
\end{tabular}
}
\caption{AUROC (\%) comparison of different methods on ImageNet-200 full-spectrum benchmarks.}
\label{table:fs}
\end{table}

\subsection{Fine-grained OOD Detection Benchmark: NABirds}
For fine-grained OOD detection, we use the NABirds \cite{van2015building} dataset, specifically designed to evaluate the model's performance in recognizing fine-grained categories.
\paragraph{Dataset Details.}The NABirds dataset contains over 48,000 images of 555 North American bird species for fine-grained bird species recognition. Each image is annotated with detailed species-level labels and metadata, presenting challenges such as high inter-species similarity and intra-species variability. Following \cite{Jin2023Mixture}, we split the dataset into three parts. In each split, a portion of the classes is used as ID data, while the holdout classes from that split are evaluated as fine OOD data. Additionally, we utilize datasets from other sources \cite{maji2013fine, krause20133d, chen2018fine} as coarse OOD data. We validated using the same implementation details as described in Appendix~\ref{sppl:impl_detail}.

\paragraph{Results.} While our main focus is on capturing subtle differences, this approach is also beneficial in more fine-grained OOD detection. We evaluate on NABirds benchmark, which consists of various bird species.  We set a portion of NABirds category as ID, the remaining species as fine-grained OOD, and entirely different benchmarks as coarse OOD datasets~\cite{Jin2023Mixture}. In this challenging setting, our method consistently improves performance across all cases, confirming its effectiveness in distinguishing fine-grained semantic differences (Table~\ref{birds}). We expect that applying more precise semantic guidance will further enhance fine-grained OOD detection performance.

\begin{table}[htbp]
\centering
\small
\begin{tabular}{@{}lcccccc@{}}
\toprule
\multicolumn{1}{c}{\multirow{2}{*}{Method}} & \multicolumn{2}{c}{split 0} & \multicolumn{2}{c}{split 1} & \multicolumn{2}{c}{split 2}  \\
\cmidrule(l){2-7} 
\multicolumn{1}{c}{} & coarse & fine & coarse & fine & coarse & fine  \\
\midrule
MSP & 92.06 & 74.57 & 90.44 & 75.16 & 90.19 & 76.73  \\
EBO & 93.55 & 72.08 & 91.05 & 74.11 & 92.21 & 72.33  \\ 
KNN & 92.65 & 73.84 & 90.76 & 74.57 & 91.74 & 74.75  \\  \midrule
\rowcolor{cyan!15} SONA (ours) & \textbf{98.97} & \textbf{77.46} & \textbf{98.88} & \textbf{78.85} & \textbf{98.78} & \textbf{79.27}  \\
\bottomrule
\end{tabular}
\caption{AUROC (\%) of fine-grained OOD detection, NABirds.}
\label{birds}
\end{table}

\section{Additional Comparison with Input Corrupting Approaches.} \label{sppl:img_edit}
\subsection{Image Editing Approach}
We also compare our method with existing Image Editing (IE) approaches as shown on Table~\ref{imageediting}. While some IE-generated samples may serve as auxiliary outliers, directly applying these conventional IE frameworks ~\cite{couairon2022diffedit, Brack2024SEGA, tsaban2023ledits,yang2023object} to OOD detection presents two significant limitations. First, samples generated during the early inversion step are nearly indistinguishable from the original ID data, which can confuse the classifier when these samples are used as OOD examples. Second, when applying target prompt-specific modifications, the resulting samples often become far outliers, lacking the necessary diversity in semantic mismatch information required for effective OOD detection. As demonstrated in the table below, simply applying existing IE methods without adaptation fails to achieve satisfactory OOD detection performance.

\begin{table}[htbp]
\centering
\caption{AUROC (\%) comparison with recent Image Editing methods}
\small 
\resizebox{0.3\columnwidth}{!}{%
\begin{tabular}{@{}lcc@{}}
\toprule
\multicolumn{1}{c}{\multirow{2}{*}{Method}}  & \multicolumn{2}{c}{AUROC (\%)} \\ \cmidrule{2-3}
 & Near-OOD & Far-OOD \\ \midrule
DiffEdit & 82.67 & 92.38 \\
SEGA & 83.21 & 92.89 \\
LEdits & 83.28 & 92.54 \\
OIR & 84.11 & 93.08 \\
\rowcolor{cyan!15}SONA (Ours) & \textbf{88.38} & \textbf{95.87} \\ \bottomrule
\end{tabular}
}\label{imageediting}
\end{table}

\subsection{Data-Centric Approach}
Our approach offers significant advantages over existing data-centric methods for OOD detection that directly corrupt input samples. Firstly, while previous methods have attempted to generate near-OOD samples, they often produce images that are either unnatural or too dissimilar from ID samples, leading to poor performance in near-OOD settings. As demonstrated in Table~\ref{datacentric}, our SONA method outperforms these simpler techniques by effectively generating outliers with subtle semantic differences. Secondly, unlike another diffusion-based outlier generation methods such as Fake-it, SONA leverages Stable diffusion model exclusively for the sampling phase, not for training. This approach significantly reduces the complexity and computational overhead, allowing us to efficiently generate high-quality outliers without the need for extensive training processes. This combination of nuanced sample generation and computational efficiency positions SONA as a robust solution for OOD detection challenges.

\begin{table}[htbp]
\centering
\small 
\resizebox{0.6\columnwidth}{!}{%
\begin{tabular}{@{}lccc@{}}
\toprule
\multicolumn{1}{c}{\multirow{2}{*}{Method}}  & \multicolumn{2}{c}{AUROC (\%)} & \multirow{2}{*}{\centering Diffusion Training} \\ \cmidrule{2-3}
 & Near-OOD & Far-OOD &  \\ \midrule
CutPaste & 77.86 & 88.18 & - \\
AutoAugOOD (from UNODE) & 81.26 & 93.31 & - \\
Fake-it & 79.34 & 89.62 & O \\
\rowcolor{cyan!15} SONA (Ours) & \textbf{87.19} & \textbf{94.52} & X \\ \bottomrule
\end{tabular}
}
\caption{AUROC (\%) comparison with recent Data-Centric methods}
\label{datacentric}
\end{table}

\section{Broader Impact.}

The SONA framework demonstrates significant potential for broad applications, particularly in settings where reliable OOD detection is crucial. In anomaly detection tasks, SONA can enhance the resilience of models when encountering novel or unexpected samples. Additionally, our methodology can contributes as a reliable near-OOD test benchmark, facilitating the advancement and evaluation of future methods focused on semantic distinctions in near-OOD scenarios.

\paragraph{A Proposal of a SONA Test Dataset.}
The SONA dataset is designed to maintain the nuisance information of the ID while introducing subtle semantic differences, making it a more refined test dataset compared to existing Near-OOD datasets. To empirically validate the effectiveness of our SONA test dataset, we evaluated the performance of well-known OOD detection methods on this dataset. As the results are presented in Table~\ref{sonatest}, when ImageNet-200 given as the ID dataset, the SONA test dataset showed significantly lower OOD detection performance compared to the exisiting Near-OOD datasets SSB-hard and NINCO. This performance degradation demonstrates that the SONA test dataset successfully generates more challenging Near-OOD samples. Furthermore, it suggests that the SONA test dataset can serve as a valuable benchmark for developing and evaluating future semantic-focused Near-OOD detection models.

\begin{table}[htbp]
\centering
\small

\begin{tabular}{@{}llcccc@{}}
\toprule
\multicolumn{2}{c}{Method}                             & SSB-hard & NINCO & SONA (ours) \\ 
\midrule
\multirow{3}{*}{Post-hoc}      & MSP                  & 80.38             & 86.29          & 66.26                \\
                               & EBO                  & 79.83             & 86.17          & 67.8                 \\
                               & KNN                  & 77.03             & 86.1           & 69.56                \\
\midrule
\multirow{1}{*}{Training w/ outliers} & OE           & 82.34             & 87.35          & 71.64                \\
\bottomrule
\end{tabular}
\caption{AUROC (\%) comparison across a Near-OOD datasets, including SONA test dataset on ImageNet-200 as the ID}
\label{sonatest}
\end{table}

\section{Challenging Cases}
We present some challenging cases, including geological formations (e.g., cliffs, lakeside, iceberg, sand dune), to demonstrate the capabilities of SONA samples. These geological formations exhibit diverse and complex textures, characterized by intricate and detailed patterns, making structural delineation challenging. For instance, cliffs feature irregular, rugged surfaces with layered compositions, while lakesides involve dynamic interactions between water and surrounding vegetation. Similarly, brain corals display highly intricate, maze-like patterns that require precise semantic identification. Our primary goal is not to achieve perfect semantic replacement but to identify and guide the semantics to generate natural corruptions. While target semantic masks and stop timesteps varied across cases, SONA consistently produced realistic outliers while maintaining contextual coherence. This highlights SONA's ability to robustly identify even challenging semantics, demonstrating its adaptability across diverse and intricate scenarios.

 \begin{figure}[htbp]
    \centering
	\includegraphics[width=0.9\linewidth]{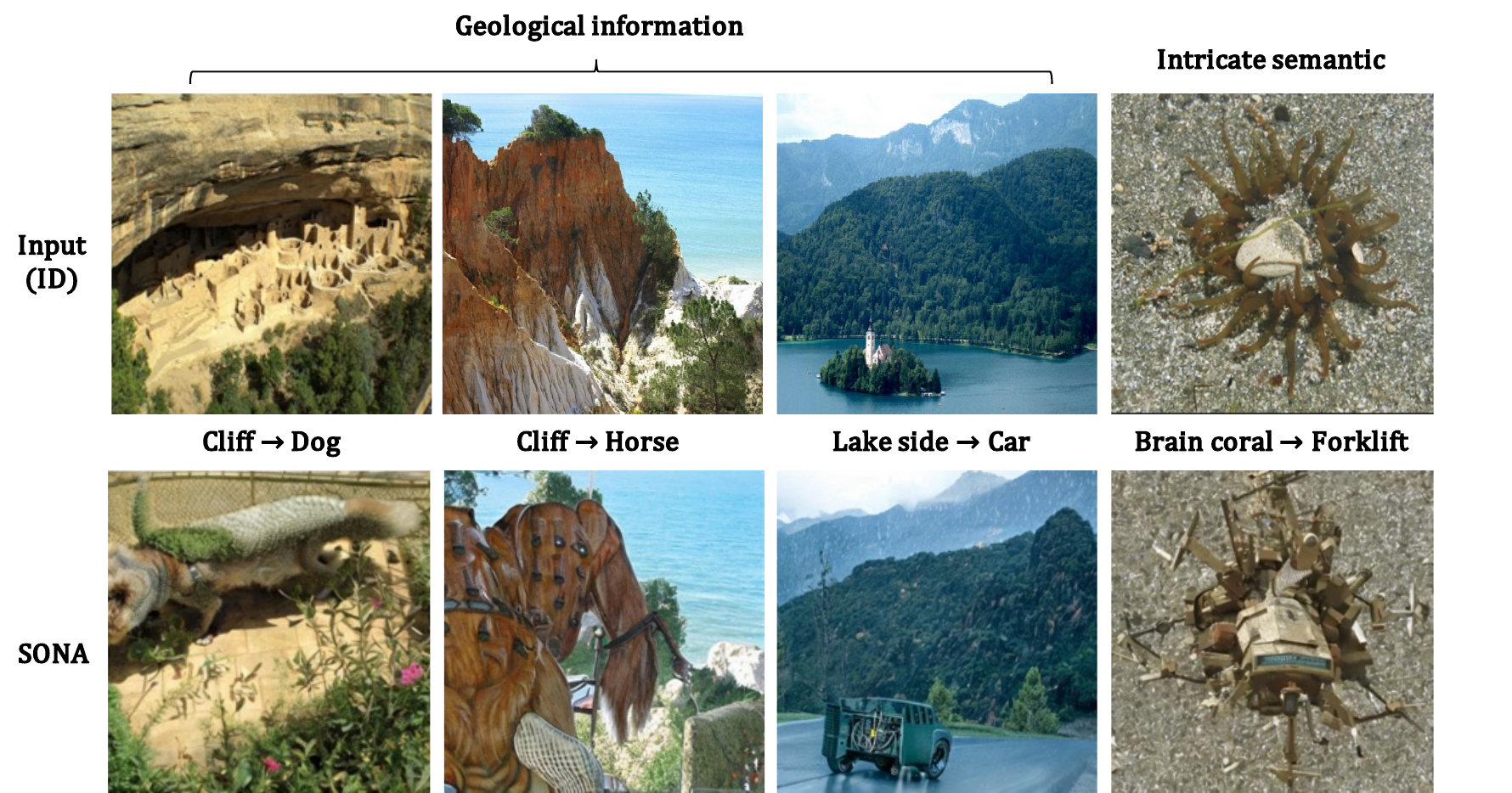}
	\caption{Challenging case of the SONA samples on ImageNet-1k dataset as the ID dataset.}
\end{figure}

\section{More Qualitative Results}

\begin{figure}[htbp]
    \centering
	\includegraphics[width=0.9\linewidth]{aaai_figures/supple/gen_sup1.pdf}
	\caption{Class-specific visual examples of the SONA samples on ImageNet-200 dataset as the ID dataset.}
\end{figure}

\begin{figure}[htbp]
    \centering
	\includegraphics[width=0.9\linewidth]{aaai_figures/supple/gen_sup2.pdf}
	\caption{Class-specific visual examples of the SONA samples on ImageNet-200 dataset as the ID dataset.}
\end{figure}